\documentclass[letterpaper, 10 pt, conference]{ieeeconf}

\IEEEoverridecommandlockouts                              

\usepackage{amsmath} 
\usepackage{amssymb}  

\title{\LARGE \bf
How to pick the domain randomization parameters for sim-to-real transfer of reinforcement learning policies?
}

\author{Quan Vuong$^{1}$, Sharad Vikram$^{2}$, Hao Su$^{3}$, Sicun Gao$^{4}$, Henrik I. Christensen$^{5}$
\thanks{$^{1}$Correspondence to qvuong@ucsd.edu}
\thanks{All authors are affiliated with the Computer Science and Engineering Department, University of California San Diego}
}

\usepackage{hyperref}
\usepackage{physics}
\usepackage{import}
\usepackage{bm}

\DeclareMathOperator*{\argmax}{argmax}

\newcommand{\cS}{\mathcal{S}}
\newcommand{\cA}{\mathcal{A}}
\newcommand{\E}[3]{\overset{#2}{\underset{#1}{\mathbb{E}}} \left[ #3 \right] }

\renewcommand{\real}{\textrm{real}}
\newcommand{\dart}{\textrm{dart}}
\newcommand{\R}{\mathbb{R}}

\begin{document}

\maketitle
\thispagestyle{empty}
\pagestyle{empty}

\section{INTRODUCTION}

Recently, reinforcement learning (RL) algorithms have demonstrated remarkable success in learning complicated  behaviors from minimally processed input (\cite{SAC, SPU, TRPO, TD3, T4DG, blackbox_RL}). However, most of this success is limited to simulation. While there are promising successes in applying RL algorithms directly on real systems (\cite{SOLAR, benchmarking_on_real_robot, autoRL, prl_rl, transfer_quadupred_robots}), their performance on more complex systems remains bottle-necked by the relative data inefficiency of RL algorithms. Domain randomization is a promising direction of research that has demonstrated impressive results using RL algorithms to control real robots (\cite{domain_rand_object_localization, domain_rand_dynamics, strategy_optim, dactyl, sim_to_sim}).

At a high level, domain randomization works by training a policy on a distribution of environmental conditions in simulation. If the environments are diverse enough, then the policy trained on this distribution will plausibly generalize to the real world. A human-specified design choice in domain randomization is the form and parameters of the distribution of simulated environments. It is unclear how to the best pick the form and parameters of this distribution and prior work uses hand-tuned distributions. This extended abstract demonstrates that the choice of the distribution plays a major role in the performance of the trained policies in the real world and that the parameter of this distribution can be optimized to maximize the performance of the trained policies in the real world.\footnote{Dockerized code to reproduce our experiments is available at \url{https://github.com/quanvuong/domain_randomization}.}
\section{Background and Notation}

In RL, the robotic learning problem is abstracted as a discrete time sequential decision making problem in a Markov decision process (MDP). An MDP is a tuple $(\cS, \cA, r, T, \gamma, \rho)$ with state space $\cS$, action space $\cA$, reward function $r: \cS \times \cA \rightarrow \mathbb{R}$, state transition function $T: \cS \times \cA \rightarrow \cS $, a discount factor $\gamma$ and a distribution over the initial state $\rho$. Given a state $s \in \cS$, a policy $\pi_\theta$ defines a distribution $\pi_\theta(.|s)$ over the action space $\cA$. $\theta$ represents the parameters of the policy, which can be linear operators (\cite{kakade_linear, recht_linear}) or the weights and biases of a deep neural network. Let $m$ denotes one specific MDP $(S^{(m)}, A^{(m)}, r^{(m)}, T^{(m)}, \gamma, \rho^{(m)})$. The performance of a policy $\pi_\theta$ with respect to the MDP parameterized by $m$ is evaluated by:
\begin{align*}
	J^{(m)}(\pi_\theta) \triangleq \E{\tau \sim \pi_\theta}{}{\sum_{t=0}^\infty \gamma^t r(s_t, a_t)}
\end{align*}
where $\tau = (s_0, a_0, r_0, s_1, \ldots)$ is a trajectory generated by using the policy $\pi_\theta$ to interact with the MDP $m$. Let $m_{\real}$ denotes the MDP representing the real world. Formally, domain randomization performs the optimization
\begin{align*}
	\theta^* = \argmax_\theta \E{m \sim p_\phi}{}{J^{(m)}(\pi_\theta)}
\end{align*}
where $p_\phi$ is a distribution over MDPs parameterized by $\phi$. $\pi_{\theta^*}$ is then used to perform the task of interest in $m_{\real}$. For example, in \cite{domain_rand_dynamics} where domain randomization was successfully used to transfer a policy trained in simulation to the real world on object pushing tasks, $\phi$ parameterizes the distribution over the masses and damping coefficients of the robot's links in addition to other environmental conditions. 

\section{Optimization of the domain randomization parameters}

At a high level, domain randomization is a technique to accomplish the general goal: ``Given a simulator, we want to use it such that when we train a policy in the simulator, the policy will perform well in the real world". We argue that this is an objective that we can optimize for directly. In prior works, the parameter $\phi$ of the distribution over MDPs is chosen by hand, presumably using domain knowledge and through trial-and-error; it is also kept fixed throughout the training process.
Prior works also assume that there is a clear demarcation between training and testing, i.e. during training in simulation, the policy does not have access to the real system. However, in practice, this assumption could be incorrect, as we may have limited or costly access to the real system.
In such scenarios, we could use 
the real system to provide some signal 
for domain randomization.

With access to the real environment $m_\real$, we can formalize domain randomization as a bilevel optimization problem:
\begin{align}
&    \argmax_\phi \quad J^{(m_{\real})} (\pi_{\theta^*(\phi)}) \label{eq:true_prob_of_interest_outer} \\
& \text{such that} \quad 	\theta^*(\phi) = \argmax_\theta \E{m \sim p_\phi}{} {J^{(m)}(\pi_\theta)}
\label{eq:true_prob_of_interest_inner}
\end{align}

To establish that this is a research direction worth pursuing, we need to demonstrate the following:

\begin{itemize}
	\item The choice of the parameter $\phi$ plays a major role in the performance of the policies in the real environment.
	\item $\phi$ can be optimized to increase the performance of the trained policies in the real environment.
\end{itemize}

We experimentally demonstrate these two points by using Cross Entropy Method (CEM) to approximately solve the outer problem (\autoref{eq:true_prob_of_interest_outer}) and Proximal Policy Optimization  (PPO) \cite{PPO} to solve the inner problem (\autoref{eq:true_prob_of_interest_inner}). The closest related work to ours is \cite{dieter_fox_sim_to_real}, which finds the simulation parameters that bring the state distribution in simulation close to the state distribution in the real world. We argue that this is only a proxy measure of the actual objective we ultimately care about and optimize for directly, i.e. the performance of the trained policy in the real environment. Other than domain randomization, other parallel research directions for sim-to-real transfer exist and have been demonstrated to be promising research areas as well (\cite{supervised_transfer_learning, sysid_karen, modulating_traj_gen, tensegrity_robot, transfer_quadupred_robots}).

\section{Algorithmic Description}

CEM is a simple iterative gradient-free stochastic optimization method. Given the decision variable $\phi$, CEM alternatives between evaluating its current value on the objective function (\autoref{eq:true_prob_of_interest_outer}) and updating $\phi$. We refer interested readers to \cite{CEM} for a more detailed description. We initialize $\phi$ with $\phi_0$, evaluate $\phi_0$ to obtain $J^{(m_{\real})} (\pi_{\theta^*(\phi_0)})$, use the evaluation result to update $\phi_0$ to obtain $\phi_1$, and so on.

\section{Experimental Settings and Results}

To demonstrate the potential of our research direction, we focus on transferring learned policies between two simulators. Specifically, we focus on transferring policies for the environments Hopper and Walker from the Dart simulator \cite{dart} to the Mujoco simulator \cite{mujoco}. Thus, $m_{\real}$ represents the parameters of the MDP in the Mujoco simulator. Transferring between these two simulators has been demonstrated to be a fruitful experimental testbed for sim-to-real studies \cite{strategy_optim}. In our setting, the MDPs in both simulators are parameterized by the masses, damping coefficients of the robot's links and the gravity constant ($\R^9$ for Hopper and $\R^{15}$ for Walker). 

$\phi$ represents the parameters of a distribution over $m$. In our experiments, $\phi$ is the mean and variance of a diagonal Gaussian distribution over the simulation parameters. The initial mean $\phi_0$ is set to $m_{\dart}$ and initial variance set to 1 for all parameters. These values are reasonable defaults for the domain randomization distribution parameters without domain knowledge or trial-and-error. CEM is then used to optimize for $\phi$.

We replicate our results for each setting over 5 different random seeds. In the Hopper environment, the performance of the policies trained with the optimized $\phi$ is on averaged $102\%$ higher than the performance of the policy trained with the initial value $\phi_0$ with a standard deviation of $48\%$ and minimum improvement of $28\%$. In the Walker environment, the performance of the policies trained with the optimized $\phi$ is on average $80\%$ higher than the performance of the policy trained with the initial value $\phi_0$ with a standard deviation of $53\%$ and minimum improvement of $19\%$. The existence of a better value for $\phi$ than $\phi_0$ shows that environment distributions chosen by hand can be improved with optimization.
Furthermore, our result is consistent with  ongoing research in domain randomization for supervised learning which demonstrated the importance of the sampling distribution for sim-to-real transfer success (\cite{learning_to_simulate, structured_domain_rand}).

\section{Future Research Directions}

\subsection{Learning complex distributions}
We assume a diagonal Gaussian sampling distribution for simplicity, but 
learning a more complex distribution
could result in a better randomized environments. For example,
deep generative modeling approaches
such as variational autoencoders (\cite{Kingma2013, Rezende14}) and
autoregressive flows (\cite{Papamakarios2017, Kingma2016}) could
be used to model complex dependencies and correlations between simulation parameters.

\subsection{Optimization techniques}
CEM was chosen to solve the outer problem (\ref{eq:true_prob_of_interest_outer}) due to its simplicity. We are interested in more advanced gradient-free optimization methods, such as CMA \cite{CMA} or Bayesian optimization \cite{bayesian_optim}. If we assume that the parameter $\phi$ is parameterized by a distribution $p_\omega$, it can be shown that $\grad_{\omega}{\underset{\phi \sim p_\omega}{E} [ J^{(m_{\real})} (\pi_{\theta^*(\phi)}) ]} = \underset{\phi \sim p_\omega}{E} [\grad_{\omega} \log p_\omega(\phi) J^{(m_{\real})} (\pi_{\theta^*(\phi)})]$ and we could apply stochastic gradient-based techniques to directly optimize for $\omega$.
We are also particularly excited about asynchronous evolutionary algorithms (AEA).
Whereas previous techniques are synchronous by nature, AEA enables simultaneously training policies in simulation and evaluating in reality, thereby making the best use of the available resources in terms of wall-clock time.

\subsection{Off-policy Reinforcement Learning}

PPO, an on-policy RL algorithm, was chosen to solve the inner problem (\ref{eq:true_prob_of_interest_inner}) due to its simplicity and speed. However, off-policy training of the policy with real world data has been demonstrated to improve the policy performance (\cite{sim_to_sim, drl_for_robotic_manipulation}). In our setting, the real world data generated to evaluate the policy at every iteration of solving the outer problem (\ref{eq:true_prob_of_interest_outer}) can be used to optimize the next policy in an off-policy fashion. Preferably, the inner problem (\ref{eq:true_prob_of_interest_inner}) is solved by an off-policy algorithm to allow for easy fine-tuning of the trained policy on real world data.

\subsection{Transferable Domain Randomization Parameters and Testing On Real Robots}

It would be of interest to understand if there exists general principles to determine the value of $\phi$ or transferable initial values for $\phi$ that works for domain randomization across a wide range of tasks and robots. This is so that the expensive problem (\autoref{eq:true_prob_of_interest_outer} and \ref{eq:true_prob_of_interest_inner}) does not have to be solved from scratch for every problem instance. Ultimately, the goal is to test our approach to domain randomization on real robots.




\section*{ACKNOWLEDGMENT}

The authors would like to acknowledge Shuang Liu for helpful discussions.

\bibliographystyle{IEEEtran}
\bibliography{IEEEabrv,IEEEexample}

\begin{thebibliography}{10}
\providecommand{\url}[1]{#1}
\csname url@rmstyle\endcsname
\providecommand{\newblock}{\relax}
\providecommand{\bibinfo}[2]{#2}
\providecommand\BIBentrySTDinterwordspacing{\spaceskip=0pt\relax}
\providecommand\BIBentryALTinterwordstretchfactor{4}
\providecommand\BIBentryALTinterwordspacing{\spaceskip=\fontdimen2\font plus
\BIBentryALTinterwordstretchfactor\fontdimen3\font minus
  \fontdimen4\font\relax}
\providecommand\BIBforeignlanguage[2]{{%
\expandafter\ifx\csname l@#1\endcsname\relax
\typeout{** WARNING: IEEEtran.bst: No hyphenation pattern has been}%
\typeout{** loaded for the language `#1'. Using the pattern for}%
\typeout{** the default language instead.}%
\else
\language=\csname l@#1\endcsname
\fi
#2}}

\bibitem{SAC}
\BIBentryALTinterwordspacing
T.~Haarnoja, A.~Zhou, P.~Abbeel, and S.~Levine, ``Soft actor-critic: Off-policy
  maximum entropy deep reinforcement learning with a stochastic actor,''
  \emph{CoRR}, vol. abs/1801.01290, 2018. [Online]. Available:
  \url{http://arxiv.org/abs/1801.01290}
\BIBentrySTDinterwordspacing

\bibitem{SPU}
\BIBentryALTinterwordspacing
Q.~H. Vuong, Y.~Zhang, and K.~W. Ross, ``Supervised policy update,''
  \emph{CoRR}, vol. abs/1805.11706, 2018. [Online]. Available:
  \url{http://arxiv.org/abs/1805.11706}
\BIBentrySTDinterwordspacing

\bibitem{TRPO}
\BIBentryALTinterwordspacing
J.~Schulman, S.~Levine, P.~Moritz, M.~I. Jordan, and P.~Abbeel, ``Trust region
  policy optimization,'' \emph{CoRR}, vol. abs/1502.05477, 2015. [Online].
  Available: \url{http://arxiv.org/abs/1502.05477}
\BIBentrySTDinterwordspacing

\bibitem{TD3}
\BIBentryALTinterwordspacing
S.~Fujimoto, H.~van Hoof, and D.~Meger, ``Addressing function approximation
  error in actor-critic methods,'' \emph{CoRR}, vol. abs/1802.09477, 2018.
  [Online]. Available: \url{http://arxiv.org/abs/1802.09477}
\BIBentrySTDinterwordspacing

\bibitem{T4DG}
\BIBentryALTinterwordspacing
G.~Barth{-}Maron, M.~W. Hoffman, D.~Budden, W.~Dabney, D.~Horgan, D.~TB,
  A.~Muldal, N.~Heess, and T.~P. Lillicrap, ``Distributed distributional
  deterministic policy gradients,'' \emph{CoRR}, vol. abs/1804.08617, 2018.
  [Online]. Available: \url{http://arxiv.org/abs/1804.08617}
\BIBentrySTDinterwordspacing

\bibitem{blackbox_RL}
\BIBentryALTinterwordspacing
K.~Choromanski, A.~Pacchiano, J.~Parker-Holder, J.~Hsu, A.~Iscen, D.~Jain, and
  V.~Sindhwani, ``When random search is not enough: Sample-efficient and
  noise-robust blackbox optimization of rl policies,'' 2019. [Online].
  Available: \url{https://arxiv.org/abs/1903.02993}
\BIBentrySTDinterwordspacing

\bibitem{SOLAR}
\BIBentryALTinterwordspacing
M.~Zhang, S.~Vikram, L.~Smith, P.~Abbeel, M.~J. Johnson, and S.~Levine,
  ``{SOLAR:} deep structured latent representations for model-based
  reinforcement learning,'' \emph{CoRR}, vol. abs/1808.09105, 2018. [Online].
  Available: \url{http://arxiv.org/abs/1808.09105}
\BIBentrySTDinterwordspacing

\bibitem{benchmarking_on_real_robot}
\BIBentryALTinterwordspacing
A.~R. Mahmood, D.~Korenkevych, G.~Vasan, W.~Ma, and J.~Bergstra, ``Benchmarking
  reinforcement learning algorithms on real-world robots,'' \emph{CoRR}, vol.
  abs/1809.07731, 2018. [Online]. Available:
  \url{http://arxiv.org/abs/1809.07731}
\BIBentrySTDinterwordspacing

\bibitem{autoRL}
H.~L. {Chiang}, A.~{Faust}, M.~{Fiser}, and A.~{Francis}, ``Learning navigation
  behaviors end-to-end with autorl,'' \emph{IEEE Robotics and Automation
  Letters}, vol.~4, no.~2, pp. 2007--2014, April 2019.

\bibitem{prl_rl}
\BIBentryALTinterwordspacing
A.~Faust, O.~Ramirez, M.~Fiser, K.~Oslund, A.~Francis, J.~Davidson, and
  L.~Tapia, ``Prm-rl: Long-range robotic navigation tasks by combining
  reinforcement learning and sampling-based planning,'' Brisbane, Australia,
  2018, pp. 5113--5120. [Online]. Available:
  \url{https://arxiv.org/abs/1710.03937}
\BIBentrySTDinterwordspacing

\bibitem{transfer_quadupred_robots}
\BIBentryALTinterwordspacing
J.~Tan, T.~Zhang, E.~Coumans, A.~Iscen, Y.~Bai, D.~Hafner, S.~Bohez, and
  V.~Vanhoucke, ``Sim-to-real: Learning agile locomotion for quadruped
  robots,'' \emph{CoRR}, vol. abs/1804.10332, 2018. [Online]. Available:
  \url{http://arxiv.org/abs/1804.10332}
\BIBentrySTDinterwordspacing

\bibitem{domain_rand_object_localization}
\BIBentryALTinterwordspacing
J.~Tobin, R.~Fong, A.~Ray, J.~Schneider, W.~Zaremba, and P.~Abbeel, ``Domain
  randomization for transferring deep neural networks from simulation to the
  real world,'' \emph{CoRR}, vol. abs/1703.06907, 2017. [Online]. Available:
  \url{http://arxiv.org/abs/1703.06907}
\BIBentrySTDinterwordspacing

\bibitem{domain_rand_dynamics}
\BIBentryALTinterwordspacing
X.~B. Peng, M.~Andrychowicz, W.~Zaremba, and P.~Abbeel, ``Sim-to-real transfer
  of robotic control with dynamics randomization,'' \emph{CoRR}, vol.
  abs/1710.06537, 2017. [Online]. Available:
  \url{http://arxiv.org/abs/1710.06537}
\BIBentrySTDinterwordspacing

\bibitem{strategy_optim}
\BIBentryALTinterwordspacing
W.~Yu, C.~K. Liu, and G.~Turk, ``Policy transfer with strategy optimization,''
  \emph{CoRR}, vol. abs/1810.05751, 2018. [Online]. Available:
  \url{http://arxiv.org/abs/1810.05751}
\BIBentrySTDinterwordspacing

\bibitem{dactyl}
\BIBentryALTinterwordspacing
OpenAI, M.~Andrychowicz, B.~Baker, M.~Chociej, R.~J{\'{o}}zefowicz, B.~McGrew,
  J.~W. Pachocki, J.~Pachocki, A.~Petron, M.~Plappert, G.~Powell, A.~Ray,
  J.~Schneider, S.~Sidor, J.~Tobin, P.~Welinder, L.~Weng, and W.~Zaremba,
  ``Learning dexterous in-hand manipulation,'' \emph{CoRR}, vol.
  abs/1808.00177, 2018. [Online]. Available:
  \url{http://arxiv.org/abs/1808.00177}
\BIBentrySTDinterwordspacing

\bibitem{sim_to_sim}
\BIBentryALTinterwordspacing
S.~James, P.~Wohlhart, M.~Kalakrishnan, D.~Kalashnikov, A.~Irpan, J.~Ibarz,
  S.~Levine, R.~Hadsell, and K.~Bousmalis, ``Sim-to-real via sim-to-sim:
  Data-efficient robotic grasping via randomized-to-canonical adaptation
  networks,'' \emph{CoRR}, vol. abs/1812.07252, 2018. [Online]. Available:
  \url{http://arxiv.org/abs/1812.07252}
\BIBentrySTDinterwordspacing

\bibitem{kakade_linear}
\BIBentryALTinterwordspacing
A.~Rajeswaran, K.~Lowrey, E.~Todorov, and S.~Kakade, ``Towards generalization
  and simplicity in continuous control,'' \emph{CoRR}, vol. abs/1703.02660,
  2017. [Online]. Available: \url{http://arxiv.org/abs/1703.02660}
\BIBentrySTDinterwordspacing

\bibitem{recht_linear}
\BIBentryALTinterwordspacing
H.~Mania, A.~Guy, and B.~Recht, ``Simple random search provides a competitive
  approach to reinforcement learning,'' \emph{CoRR}, vol. abs/1803.07055, 2018.
  [Online]. Available: \url{http://arxiv.org/abs/1803.07055}
\BIBentrySTDinterwordspacing

\bibitem{PPO}
\BIBentryALTinterwordspacing
J.~Schulman, F.~Wolski, P.~Dhariwal, A.~Radford, and O.~Klimov, ``Proximal
  policy optimization algorithms,'' \emph{CoRR}, vol. abs/1707.06347, 2017.
  [Online]. Available: \url{http://arxiv.org/abs/1707.06347}
\BIBentrySTDinterwordspacing

\bibitem{dieter_fox_sim_to_real}
\BIBentryALTinterwordspacing
Y.~Chebotar, A.~Handa, V.~Makoviychuk, M.~Macklin, J.~Issac, N.~D. Ratliff, and
  D.~Fox, ``Closing the sim-to-real loop: Adapting simulation randomization
  with real world experience,'' \emph{CoRR}, vol. abs/1810.05687, 2018.
  [Online]. Available: \url{http://arxiv.org/abs/1810.05687}
\BIBentrySTDinterwordspacing

\bibitem{supervised_transfer_learning}
\BIBentryALTinterwordspacing
J.~Hwangbo, J.~Lee, A.~Dosovitskiy, D.~Bellicoso, V.~Tsounis, V.~Koltun, and
  M.~Hutter, ``Learning agile and dynamic motor skills for legged robots,''
  \emph{CoRR}, vol. abs/1901.08652, 2019. [Online]. Available:
  \url{http://arxiv.org/abs/1901.08652}
\BIBentrySTDinterwordspacing

\bibitem{sysid_karen}
\BIBentryALTinterwordspacing
W.~Yu, V.~C. Kumar, G.~Turk, and C.~K. Liu, ``Sim-to-real transfer for biped
  locomotion,'' 2019. [Online]. Available:
  \url{https://arxiv.org/abs/1903.01390}
\BIBentrySTDinterwordspacing

\bibitem{modulating_traj_gen}
\BIBentryALTinterwordspacing
A.~Iscen, K.~Caluwaerts, J.~Tan, T.~Zhang, E.~Coumans, V.~Sindhwani, and
  V.~Vanhoucke, ``Policies modulating trajectory generators,'' in
  \emph{Proceedings of The 2nd Conference on Robot Learning}, ser. Proceedings
  of Machine Learning Research, A.~Billard, A.~Dragan, J.~Peters, and
  J.~Morimoto, Eds., vol.~87.\hskip 1em plus 0.5em minus 0.4em\relax PMLR,
  29--31 Oct 2018, pp. 916--926. [Online]. Available:
  \url{http://proceedings.mlr.press/v87/iscen18a.html}
\BIBentrySTDinterwordspacing

\bibitem{tensegrity_robot}
M.~{Zhang}, X.~{Geng}, J.~{Bruce}, K.~{Caluwaerts}, M.~{Vespignani},
  V.~{SunSpiral}, P.~{Abbeel}, and S.~{Levine}, ``Deep reinforcement learning
  for tensegrity robot locomotion,'' in \emph{2017 IEEE International
  Conference on Robotics and Automation (ICRA)}, May 2017, pp. 634--641.

\bibitem{CEM}
\BIBentryALTinterwordspacing
P.-T. de~Boer, D.~P. Kroese, S.~Mannor, and R.~Y. Rubinstein, ``A tutorial on
  the cross-entropy method.'' [Online]. Available:
  \url{http://web.mit.edu/6.454/www/www_fall_2003/gew/CEtutorial.pdf}
\BIBentrySTDinterwordspacing

\bibitem{dart}
\BIBentryALTinterwordspacing
J.~Lee1, M.~X. Grey, S.~Ha, T.~Kunz, S.~Jain, Y.~Ye, S.~S. Srinivasa1,
  M.~Stilman, , and C.~K. Liu, ``Dart: Dynamic animation and robotics
  toolkit,'' 2018. [Online]. Available:
  \url{https://personalrobotics.cs.washington.edu/publications/lee2018dart.pdf}
\BIBentrySTDinterwordspacing

\bibitem{mujoco}
\BIBentryALTinterwordspacing
E.~Todorov, T.~Erez, and Y.~Tassa, ``Mujoco: A physics engine for model-based
  control.'' [Online]. Available:
  \url{https://homes.cs.washington.edu/~todorov/papers/TodorovIROS12.pdf}
\BIBentrySTDinterwordspacing

\bibitem{learning_to_simulate}
\BIBentryALTinterwordspacing
N.~Ruiz, S.~Schulter, and M.~Chandraker, ``Learning to simulate,'' \emph{CoRR},
  vol. abs/1810.02513, 2018. [Online]. Available:
  \url{http://arxiv.org/abs/1810.02513}
\BIBentrySTDinterwordspacing

\bibitem{structured_domain_rand}
\BIBentryALTinterwordspacing
A.~Prakash, S.~Boochoon, M.~Brophy, D.~Acuna, E.~Cameracci, G.~State,
  O.~Shapira, and S.~Birchfield, ``Structured domain randomization: Bridging
  the reality gap by context-aware synthetic data,'' \emph{CoRR}, vol.
  abs/1810.10093, 2018. [Online]. Available:
  \url{http://arxiv.org/abs/1810.10093}
\BIBentrySTDinterwordspacing

\bibitem{Kingma2013}
\BIBentryALTinterwordspacing
D.~P. Kingma and M.~Welling, ``{Auto-Encoding Variational Bayes},'' 12 2013.
  [Online]. Available: \url{http://arxiv.org/abs/1312.6114}
\BIBentrySTDinterwordspacing

\bibitem{Rezende14}
D.~J. Rezende, S.~Mohamed, and D.~Wierstra, ``Stochastic backpropagation and
  approximate inference in deep generative models,'' in \emph{Proceedings of
  the 31st International Conference on Machine Learning}, ser. Proceedings of
  Machine Learning Research, E.~P. Xing and T.~Jebara, Eds., vol.~32,
  no.~2.\hskip 1em plus 0.5em minus 0.4em\relax PMLR, 22--24 Jun 2014, pp.
  1278--1286.

\bibitem{Papamakarios2017}
G.~Papamakarios, T.~Pavlakou, and I.~Murray, ``Masked autoregressive flow for
  density estimation,'' in \emph{Advances in Neural Information Processing
  Systems}, 2017, pp. 2338--2347.

\bibitem{Kingma2016}
D.~P. Kingma, T.~Salimans, R.~Jozefowicz, X.~Chen, I.~Sutskever, and
  M.~Welling, ``Improved variational inference with inverse autoregressive
  flow,'' in \emph{Advances in neural information processing systems}, 2016,
  pp. 4743--4751.

\bibitem{CMA}
\BIBentryALTinterwordspacing
I.~Loshchilov and F.~Hutter, ``Cma-es for hyperparameter optimization of deep
  neural networks,'' \emph{CoRR}, vol. abs/1604.07269, 2016. [Online].
  Available: \url{http://arxiv.org/abs/1604.07269}
\BIBentrySTDinterwordspacing

\bibitem{bayesian_optim}
\BIBentryALTinterwordspacing
K.~Kandasamy, K.~R. Vysyaraju, W.~Neiswanger, B.~Paria, C.~R. Collins,
  J.~Schneider, B.~Poczos, and E.~P. Xing, ``Tuning hyperparameters without
  grad students: Scalable and robust bayesian optimisation with dragonfly,''
  \emph{CoRR}, 2019. [Online]. Available:
  \url{https://arxiv.org/abs/1903.06694}
\BIBentrySTDinterwordspacing

\bibitem{drl_for_robotic_manipulation}
\BIBentryALTinterwordspacing
S.~Gu, E.~Holly, T.~P. Lillicrap, and S.~Levine, ``Deep reinforcement learning
  for robotic manipulation,'' \emph{CoRR}, vol. abs/1610.00633, 2016. [Online].
  Available: \url{http://arxiv.org/abs/1610.00633}
\BIBentrySTDinterwordspacing

\end{thebibliography}

\end{document}